\documentclass{article}
\usepackage{float}
\usepackage{microtype}
\usepackage{graphicx}
\usepackage{subfigure}
\usepackage{makecell}
\usepackage{booktabs} 

\usepackage{hyperref}

\usepackage[accepted]{icml2021}

\icmltitlerunning{Combining Counterfactuals With Shapley Values To Explain Image Models}

\begin{document}

\twocolumn[
\icmltitle{Combining Counterfactuals With Shapley Values To Explain Image Models}



\icmlsetsymbol{equal}{*}

\begin{icmlauthorlist}
\icmlauthor{Aditya Lahiri}{to}
\icmlauthor{Kamran Alipour}{to}
\icmlauthor{Ehsan Adeli}{sf}
\icmlauthor{Babak Salimi}{to}

\end{icmlauthorlist}

\icmlaffiliation{to}{University Of California, San Diego}
\icmlaffiliation{sf}{Department of Computer Science, Stanford University}

\icmlcorrespondingauthor{Aditya Lahiri}{adlahiri@ucsd.edu}

\icmlkeywords{Machine Learning, ICML}

\vskip 0.3in
]



\printAffiliationsAndNotice{}  

\begin{abstract}
 With the widespread use of sophisticated machine learning models in sensitive applications, understanding their decision-making has become an essential task. Models trained on tabular data have witnessed significant progress in explanations of their underlying decision making processes by virtue of having a small number of discrete features. 
However, applying these methods to high-dimensional inputs such as images is not a trivial task. Images are composed of pixels at an atomic level and do not carry any interpretability by themselves.  
In this work, we seek to use annotated high-level interpretable features of images to provide explanations. 
We leverage the Shapley value framework from Game Theory, which has garnered wide acceptance in general XAI problems. 
By developing a pipeline to generate counterfactuals and subsequently using it to estimate Shapley values, we obtain contrastive and interpretable explanations with strong axiomatic guarantees.
\end{abstract}

\section{Introduction}

Understanding the decision-making rationale behind complicated black-box models has emerged as one of the most critical tasks in the greater overarching goal of making AI systems transparent and trustworthy. Amidst several proposed methods, the well-studied concept of Shapley values~\cite{winter2002shapley} from Game Theory literature has emerged as a principled framework to obtain feature attributions as explanations. By virtue of their strong axiomatic guarantees, they lend themselves favorably to the task of distributing model outputs fairly among the different input features. However, one of the major issues in using Shapley values for model explanation is the computation time~\cite{van2021tractability}. It grows exponentially with the number of features involved. This problem is particularly exacerbated in end-to-end models that operate directly on images since, usually, the features in images are defined on a pixel level. This results in thousands of features. In addition, 
the individual pixel values are not interpretable to humans, and therefore any attribution attached to individual pixels does not help with high-level model understanding. It has also been shown that humans understand and explain things by comparison and contrast~\cite{de2017people}. On their own, feature attribution numbers obtained by Shapley values do not allow for this contrastive notion and lack a reference or visual aid to compare to. This has been an essential barrier for generating explanations based on  
Shapley values ~\cite{kumar2020problems}. 

To close the gap, we propose an approach that incorporates high-level interpretable features and employs generative models to produce counterfactual images corresponding to specific changes in the interpretable features. These counterfactuals are then used to compute Shapley values which explain the difference in prediction scores between the original and counterfactual image. This process enables us to understand model behavior using the contrastive explanations (w.r.t. Shaply values) provided for arbitrary input images.

\begin{figure*}[t]
\begin{center}
\includegraphics[width=\linewidth]{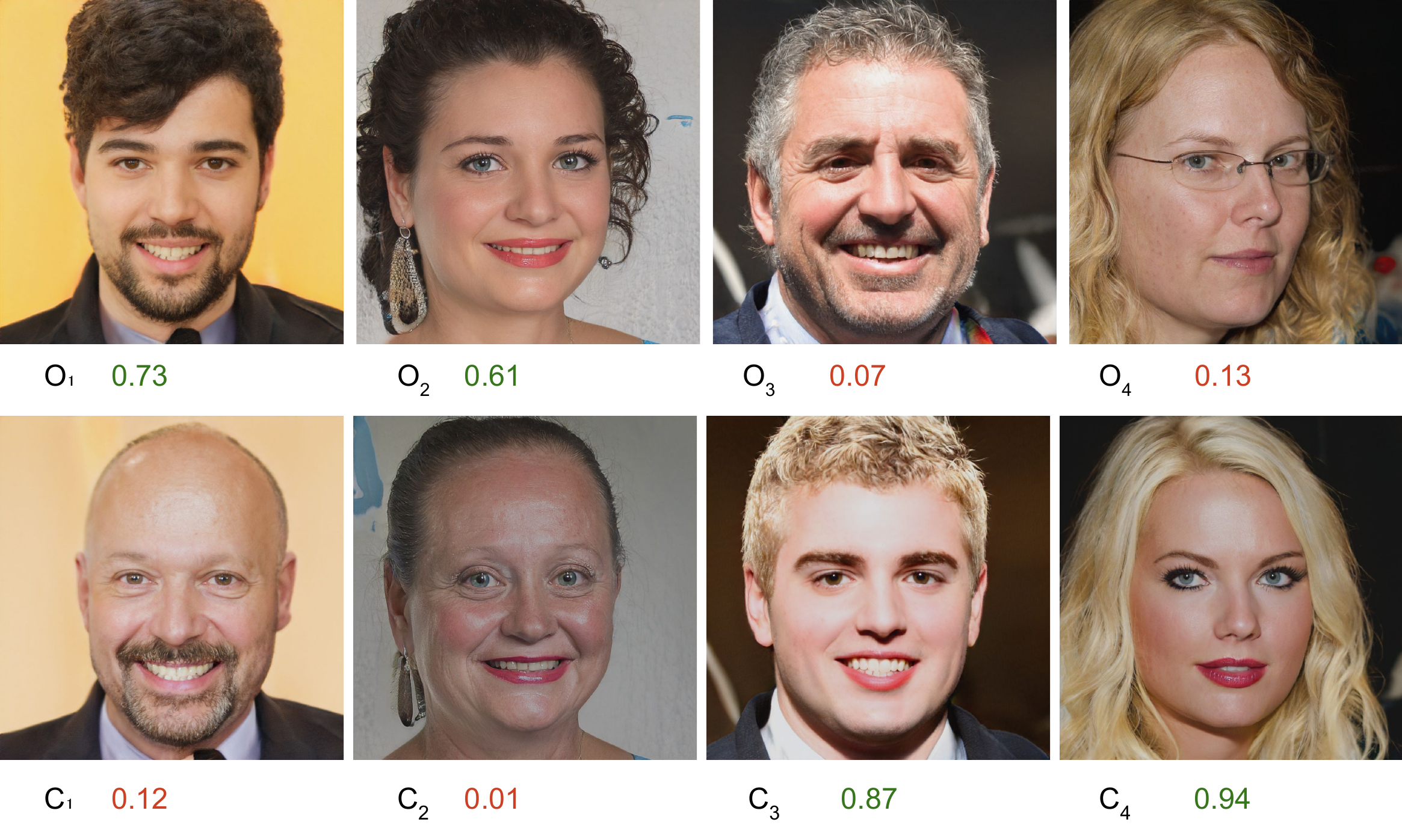}
\end{center}
\caption{
Examples of original images and their corresponding counterfactual images. The values below the images represent the model's  prediction for Attractiveness. We generate the counterfactual images according to known specifications. For instance, $C_1$ was obtained from $O_1$ by increasing baldness and maleness while decreasing youngness, heavy makeup, and blond hair. Similarly, $C_4$ was generated from $O_4$ by increasing youngness, blond hair, heavy makeup, and decreasing maleness and baldness. We use Shapley values to explain the difference between the original images and their counterfactual images in terms of the modified interpretable attributes in Table \ref{table:shapley_values}.
}
\label{fig:exp}
\end{figure*}
\section{Related Work}
Shapley values have been widely used in the context of Explainable AI (XAI) to provide feature attributions as explanations~\cite{lundberg2017unified,datta2016algorithmic}. However, since computing Shapley values is intractable~\cite{van2021tractability}, various approximations have been used~\cite{sundararajan2020many}. For models that work on images as inputs, popular techniques include aggregating neighboring pixels to form sub-pixels~\cite{ribeiro2016should} to be more interpretabile. Other works require gradients to obtain pixel-level explanations and try to compute them efficiently~\cite{li2020efficient}. We propose a model-agnostic approach that generates explanations in terms of a limited number of high-level human interpretable features. 

Significant work has gone in to producing contrastive explanations~\cite{kommiya2021towards,galhotra2021explaining}. These have primarily been shown to work on structured tabular data. Counterfactual image generation has also been an active space of research~\cite{nemirovsky2020countergan,samangouei2018explaingan}. Often they require re-training a generative model~\cite{dash2022evaluating}, which is expensive or are specific to the choice of generative model \cite{abdal2021styleflow}. We provide a minimal and scalable training process by utilizing only pretrained GANs for generating counterfactual images. 


Recent work by \cite{de2020human} comes closest to our work and seeks to use Shapley values to explain models trained on high dimensional data. They use classical computer vision techniques to generate interpretable features. In contrast, we use generative models to produce counterfactual images to compute Shapley values efficiently while also providing contrastive explanations.
\section{Background}
\subsection{Contrastive Counterfactuals \& Generative Models}
 Contrastive counterfactuals have been the building blocks of ideas in philosophy and cognition that guide people's understanding and dictate how we explain things to one another ~\cite{de2017people} and have been argued to be central to explainable AI ~\cite{miller2019explanation}.
 We are specifically interested in the implementation of this framework in the image classification problem in order to allow us to generate counterfactual images at will such that they increase or decrease the presence of a set of interpretable feature attributes.
Generative models are vastly popular in different fields of AI, and their recent advancements in creating realistic images have made them a viable approach to produce a latent representation of an image dataset. In our experiments, we utilize StyleGAN2~\cite{karras2020analyzing} as a state-of-the-art generative model which can be used to generate high resolution and realistic images in different domains. StyleGAN feeds the latent variable into a mapping network that transforms it into an intermediate latent variable. Aside from its ability to produce styles, this transformation also provides the intermediary latent space as a more regulated domain to learn and traverse through interpretable attributes. We use the manipulation of this latent space to provide us with realistic counterfactual images according to our specifications.
\subsection{Shapley Values}
Shapley value is a concept from Game Theory that provides a unique solution to fairly allocating the total payout from a game to its individual players. A coalition is a set of players playing the game. A grand coalition contains all the players, while an empty coalition contains none. A value function $v(S)$ provides the scores obtained by the coalition $S$ on playing the game. The score obtained by playing the game is called the payoff. \\
In the context of XAI, Shapley values have been used as a means to obtain feature contribution. The input features are the players in this game of obtaining predictions from the model. The prediction is then fairly distributed among the input features. For a coalition set $S$, the following formula provides the Shapley value $\phi$ for feature $i$:
\begin{equation}
   \phi_i(v)=\sum_{S \subset N \symbol{92} {i}} \frac{|S|!(n-|S|-1)!}{n!}(v(S \cup {i})-v(S)) 
\end{equation}
where $n=|N|$ is the total number of input features. This is the weighted sum of marginal contribution of feature $i$ across all possible coalitions that do not contain the feature $i$. Shapley values satisfy desirable axioms that make them make a good choice for generating feature contributions. We describe 3 of those axioms now-\\
\textbf{Null}  : The null axioms states that if a feature that does not change the output  when added to any coalition, it will get a Shapley value of zero.\\
 \textbf{Efficiency} : The efficiency property states that the sum of Shapley values is equal to the difference between the prediction obtained by the grand coalition and empty coalition.\\
 \textbf{Symmetry}: If two features behave in the same way across all possible coalitions, then their Shapley values will end up being the same.

 The grand coalition is usually the entire instance to be explained, while the null coalition is an average instance or an instance composed entirely of default replacements for missing values for each feature(since they are all out of the coalition in the null coalition). The Shapley values are then used to decompose the difference between an instance's prediction and the average prediction of the model to the individual features of the instance through the efficiency axiom.
Since we iterate over all possible subsets of features that do not contain the \emph{i}th feature, computing Shapley values in exponential in the number of features. This is particularly problematic in case of images, where we have large numbers of pixels as input features. For partial coalitions, we need to compute the model output given only the members in the coalition, which results in partially formed instances. This is usually done by either marginalizing ~\cite{datta2016algorithmic} or conditioning ~\cite{lundberg2017unified}. However, both of these techniques are known to have issues~\cite{kumar2020problems}. Further, this vanilla Shapley value-based decomposition explains away the difference between the prediction of the input instance and the model prediction on an "average" instance. This average model prediction does not always correspond to a sensible input~\cite{sundararajan2020many}. It has also been shown that standard feature contributions in isolation do not help humans reason as well as contrastive explanations do~\cite{de2017people}. We aim to overcome these issues through our modified contrastive formulation of Shapley values using high-level interpretable input features as players.


\section{Methodology}
\subsection{Generating Counterfactual Images}
We use a Shift Predictor model to obtain counterfactual images according to our needs. A shift predictor model is an MLP model that takes latent variable of an image from a generative model \(G\) and generates the latent variable for its counterfactual based on the attributes produced by a classifier. For a generative model \(G: \mathcal{R}^d\rightarrow \mathcal{R}^n\) having a latent space with dimension \(d\) and a classifier \(C: \mathcal{R}^n \rightarrow \mathcal{R}^m\) that predicts \(m\) attributes, we define our shift predictor as \(M(\textbf{z},\hat{\textbf{y}}): \mathcal{R}^d \times \mathcal{R}^m\rightarrow \mathcal{R}^d\), where  \(\textbf{z} \in \mathcal{R}^d\) is the latent variable for the input image and \(\hat{\textbf{y}}\) denotes the attributes for the intended counterfactuals. During training, shift predictor learns the directions in the latent space of \(G\) that correspond to changes in the attributes predicted by the classifier. Without need for any manual labeling, the training procedure only requires the latent variables of images from \(G\) to input the shift predictor and supervise it with the labels generated by the classifier.

During training, shift predictor learns to produce a counterfactual latent variable that satisfies any combination of attributes defined by \(\hat{\textbf{y}}\). In other words, if the classifier predicts a set of attributes \(\mathcal{A} = \{A_1, A_2, ..., A_m\}\), shift predictor provides a counterfactual latent variable compatible with any selected subset of attributes \(\bar{\mathcal{A}}\):
\begin{equation}
  \hat{\textbf{z}} = M(\textbf{z},\{A_i=\hat{a}_i\ | A_i \in \bar{\mathcal{A}}\} ).
  \label{eq:ShiftPredictorInference}
\end{equation}
Under proper training, the shift predictor is an approximation of latent variable distribution conditioned by the subset of attributes \(\bar{\mathcal{A}}\) :
\begin{equation}
  \hat{\textbf{z}} \sim P(\textbf{z}|\{A_i=\hat{a}_i\ | A_i \in \bar{\mathcal{A}}\} ),\ \bar{\mathcal{A}} \subseteq \mathcal{A}.
  \label{eq:zhatDistribution}
\end{equation}
 The loss function in the training process pursues two objectives: 1) minimizing the error in prediction of attributes \(\bar{\mathcal{A}}\) for the counterfactual image, 2) assuring a level of faithfulness of the generated counterfactual to the original image. The attribute loss \(\mathcal{L}_{a}\) is defined as a cross entropy between the conditioned attributes and the attributes predicted by the classifier. During training, the conditioned attributes \(\bar{\mathcal{A}}\) are distinguished from unset attributes so the loss will be only calculated for them. On the other hand, the faithfulness loss \(\mathcal{L}_{f} \) is calculated as the normal distance between the original latent variables and their counterfactuals. The overall loss in the training process is defined as a combination of these two losses with a faithfulness factor \(\gamma\) which defines a balance between attribute accuracy of counterfactuals and their faithfulness to the original input:
 \begin{equation}
   \mathcal{L} = \mathcal{L}_a + \gamma \mathcal{L}_f = \sum_{A_i \in \bar{\mathcal{A}}}
   -\hat{\textbf{y}}_i log(\textbf{y}_i) + \gamma ||\hat{\textbf{z}} - \textbf{z}||
  \label{eq:FullLoss}
\end{equation}

\subsection{Contrastive Explanations Using Shapley Values}
The efficiency property of Shapley values allows us to decompose the score difference between the game played by the grand coalition and the empty coalition among all the players. We define players in terms of interpretable attributes of the images. In  images of human faces, these attributes could be attributes like hair, makeup, etc. We work in this interpretable space instead of pixel-level features. We define the empty coalition as the set of all zeros for these attributes. It corresponds to their natural or default values that they take on in the image. However, we can define the grand coalition by either increasing the attributes or decreasing them. We use the shift predictor described above to obtain images corresponding to this increase or decrease in the specific attribute of the image. For instance, we can define the grand coalition to be +1 for all interpretable features, which would correspond to an image that has been forced to increase the presence of all attributes in itself. This provides us with 2 images to compare - the original image corresponding to the empty coalition and the counterfactual image with a set of increased or reduced attributes($+1 \uparrow,-1 \downarrow$). We use Shapley values to decompose the difference in prediction obtained for these two images into their interpretable features. Our shift predictor can take any arbitrary vector specifying the direction of change of attributes and return a counterfactual image. This overcomes issues found in other traditional Shapley value-based frameworks of missing data for members out of coalition. For partial coalitions, we set the values for members out of the coalition to be $0$. This instructs the shift predictor to not make any changes to that attribute. Hence, we directly compute value functions for any coalition using this pipeline. In this setup, we gain a contrastive notion and no longer compare to an "average" instance. Instead, we have two specific images- the original and its counterfactual, and we explain the difference in their predictions Shapley values.


\section{Experiments}








\begin{table}[t]
\caption{Shapley Value based contributions explaining the difference in predictions between each pair of original and counterfactual images from Figure \ref{fig:exp}. Each element is of the form $( \uparrow/ \downarrow,attribution)$ where the  $\uparrow$/ $\downarrow$ indicates whether that specific feature in the original image was increased or decreased to produce the counterfactual.}
\label{table:shapley_values}
\vskip 0.15in
\begin{center}
\begin{small}
\begin{sc}
\scalebox{0.9}{
\begin{tabular}{ |c|c|c|c|c|c| } 
\hline
Image & Young & \makecell{Heavy\\Makeup} & \makecell{Blond\\Hair} & Bald & Male\\
\hline
1 & $\downarrow$ -0.28 & $\downarrow$ -0.02 & $\downarrow$ -0.03 & $\uparrow$ -0.34 &  $\uparrow$ 0.07 \\

2 & $\downarrow$ -0.20 & $\downarrow$ -0.07 & $\downarrow$ -0.03 &  $\uparrow$ -0.23 & $\downarrow$ -0.04\\ 

3 &  $\uparrow$ 0.23 &  $\uparrow$ 0.37 &  $\uparrow$ 0.15 & $\downarrow$ 0.10 &  $\uparrow$ -0.05 \\  

4 &  $\uparrow$ 0.16 &  $\uparrow$ 0.18 &  $\uparrow$ 0.13 & $\downarrow$ 0.23 & $\downarrow$ 0.09\\
\hline

\end{tabular}}
\end{sc}
\end{small}
\end{center}
\vskip -0.4in
\end{table}
We run experiments to explain a classifier that is trained on face images. We have annotations for the set of interpretable attributes \(A\) that we will use to explain the model's behavior. We train a multi-task classifier built on top of a pretrained VGG~cite backbone to predict the set of interpretable attributes \(A\) for new unseen images. The CelebA dataset\cite{liu2015faceattributes} is used as the training set and provides 39 binary attributes, including attractiveness which we use as the target output \(Y\) for any black-box classifier of choice. As the set of interpretable explanatory attributes(\(A\)), we choose five other labels: blonde hair, heavy makeup, baldness, youngness, and maleness. We model attractiveness as the positive class $(y = 1.0)$ and unattractiveness as the negative class $(\hat{y} = 0.0)$ for the classifier to predict. The black-box classifier is built on a pretrained VGG backbone. We train our shift predictor model, as described earlier, with a faithfulness $\gamma$ value of $0.09$. We pass the images through the multi-task classifier to obtain the attribute values and through the black-box classifier to obtain the target label.\\
We sample images randomly from the StyleGan2 Generative Model and pass them through the black-box model. For every instance, we generate a counterfactual image through the shift predictor by defining our grand coalition in terms of increase or decrease of feature attributes. We obtain Shapley values-based contributions that explain away the difference in prediction between the original image(empty coalition) and the counterfactual image(grand coalition). Both the original and contrastive images along with the Shapley values of the interpretable features are given as explanations. We report the Shapley value-based attributions for images in Figure \ref{fig:exp} in Table \ref{table:shapley_values}.  The original images have their default attributes, while counterfactual images are generated through modifying the interpretable images as shown by $\uparrow$ and  $\downarrow$ signs for the increase and decrease of attributes, respectively, as listed in Table \ref{table:shapley_values}. For example, we can observe that for the original image $O_1$, we generate the counterfactual image $C_i$ by increasing maleness and baldness while decreasing youngness, heavy makeup, and blond hair. When this is done, the attractiveness score drops from $0.73$ to $0.12$. This drop is mainly due to increasing baldness which accounts for $0.34$ of the $0.61$ difference in attractiveness, while decreasing youngness(making the individual old) contributed to $0.28$ of the total drop. Similarly, we can look at other pairs of original images and counterfactuals, in conjunction with the assigned Shapley value-based attributions to understand how the model behaves.


\section{Future Work}
As next steps, we want to make the shift predictor causal by incorporating causal graphs in order to avoid modifications induced by spurious correlations among related input attributes. We also want to extend this to a systematic, automated technique to present users with a set of diverse, representative samples and their counterfactuals, with corresponding contrastive explanations to help them easily understand the model globally at a glance.


\bibliography{main}
\bibliographystyle{icml2021}
\end{document}